# Masking Strategies for Image Manifolds

Hamid Dadkhahi and Marco F. Duarte, *Senior Member, IEEE*

*Abstract*—We consider the problem of selecting an optimal mask for an image manifold, i.e., choosing a subset of the pixels of the image that preserves the manifold's geometric structure present in the original data. Such masking implements a form of compressive sensing through emerging imaging sensor platforms for which the power expense grows with the number of pixels acquired. Our goal is for the manifold learned from masked images to resemble its full image counterpart as closely as possible. More precisely, we show that one can indeed accurately learn an image manifold without having to consider a large majority of the image pixels. In doing so, we consider two masking methods that preserve the local and global geometric structure of the manifold, respectively. In each case, the process of finding the optimal masking pattern can be cast as a binary integer program, which is computationally expensive but can be approximated by a fast greedy algorithm. Numerical experiments show that the relevant manifold structure is preserved through the data-dependent masking process, even for modest mask sizes.

## I. INTRODUCTION

RECENT advances in sensing technology have enabled a massive increase in the dimensionality of data captured from digital sensing systems. Naturally, the high dimensionality of the data affects various stages of the digital systems, from data acquisition to processing and analysis. To meet communication, computation, and storage constraints, in many applications one seeks a low-dimensional embedding of the high-dimensional data that shrinks the size of the data representation while retaining the information we are interested in capturing. This problem of *dimensionality reduction* has attracted significant attention in the signal processing and machine learning communities.

The traditional method for dimensionality reduction is *principal component analysis* (PCA) [3], [4], which successfully captures the structure of datasets well approximated by a linear subspace. However, in many parameter estimation problems, the data can be best modeled by a *nonlinear manifold* whose geometry cannot be captured by PCA. Manifolds are low-dimensional geometric structures that reside in a high-dimensional ambient space despite possessing merely a few degrees of freedom. Manifold models are a good match for datasets associated with a physical system or event governed by a few continuous-valued parameters. Once the manifold model is formulated, any point on the manifold can be essentially represented by a low-dimensional parameter vector. *Manifold learning methods* aim to obtain a suitable nonlinear embedding into a low-dimensional space that preserves the geometric structure present in the higher-dimensional data. In general, manifold learning methods can be subdivided into two main categories: ($i$) techniques that attempt to preserve *global* geometry of the original data in the low-dimensional representation (e.g., Isomap [5] as well as [6]), and ($ii$) techniques that attempt to preserve *local* geometry of the original data in the low-dimensional representation (e.g., Locally Linear Embedding (LLE) [7] and others [8], [9]).

For high-dimensional data, the process of data acquisition followed by a dimensionality reduction method is inherently wasteful, since we are often not interested in obtaining the full-length representation of the data. This issue has been addressed by *compressive sensing*, a technique to simultaneously acquire and reduce the dimensionality of sparse signals in a randomized fashion [10], [11]. As an extension of compressive sensing, the use of *random projections* for linear embedding of nonlinear manifold datasets has been proposed [12]–[16], where the high-dimensional data is mapped to a random subspace of lower (but sufficiently high) dimensionality. As a result, the pairwise distances between data points are preserved with high probability.

Compressive sensing provides a good match to the requirements of cyber-physical systems, where power constraints are paramount. In such applications, one wishes to reduce the size of the representation of the data to be processed, often by applying standard compression algorithms. For instance, a fundamental challenge in the design of computational eyeglasses for gaze tracking is addressing stringent resource constraints on data acquisition and processing that include sensing fidelity and energy budget, in order to meet lifetime and size design targets [17]. Prior work in the area of compressive imaging has considered the design of linear embeddings that allow for data processing directly from lower-dimensional representation, with a particular emphasis in imaging [12]–[14], [18], [19]. However, while the aforementioned embeddings may reduce the computational and communication demands, they do not reduce the power consumption burden of data acquisition. This is because they require all image pixels to be sensed, and so they cannot be implemented more efficiently than standard acquisition.

A recent example of a power-efficient imaging architecture employs a sensor that can significantly reduce the power consumption of sensing by allowing pixel-level

This work was supported by NSF Grant IIS-1239341. Portions of this work appeared at the IEEE Statistical Signal Processing Workshop (SSP), 2014 [1] and the IEEE International Conference on Acoustics, Speech, and Signal Processing (ICASSP), 2015 [2] .

H. Dadkhahi and M. F. Duarte are with the Department of Electrical and Computer Engineering, University of Massachusetts, Amherst, MA, 01003. E-mail: {hdadkhahi,mduarte}@ecs.umass.edu.

control of the image acquisition process [20]; the power consumption of imaging grows with the number of pixels to be acquired using the array. In order to incorporate such an architecture into compressive imaging and enable the promised savings in power, we need to devise new mask selection approaches governed by the same principle of preservation of relevant image data as existing work in embedding design. Thus, it is now possible to meet stringent power and communication requirements by designing *data-dependent image masking schemes* that reduce the number of pixels involved in acquisition while, like the aforementioned linear embeddings, preserving the information of interest. The selection of a masking pattern is ideally driven by knowledge of the data model that captures the relevant information in the data, such as a nonlinear manifold model for images controlled by a few degrees of freedom.

Previous work on linear dimensionality reduction for manifolds does not address the highly constrained (masking) setting that is motivated by our application. Feature selection methods can be applied in imaging settings directly on the image pixels to obtain masks similar to those we desire [21]–[24]. Nonetheless, we note that our proposed masking approaches are designed specifically for nonlinear manifold learning, in contrast to the state of the art in feature selection.

In this paper, we consider the problem of designing masking patterns that preserve the geometric structure of a high-dimensional dataset modeled as a nonlinear manifold. The preservation of this structure through the masking is relevant to preserve the performance of manifold learning. Note that in terms of linear embeddings, masking schemes may be described as a restriction to embeddings where the projection directions are required to correspond to canonical vectors. We consider two types of geometric structure to be preserved: global structure (considered by algorithms such as Isomap) and local structure (considered by algorithms such as LLE). We test our proposed masking algorithms using Isomap and LLE because they are the most widely known and used manifold learning methods in the literature, each belonging to a different category of manifold learning methods, i.e., global and local manifold learning methods [25], respectively. Additionally, for the motivating example on the eyeglasses dataset, LLE has been used in the past to obtain eye gaze location estimates, cf. [26].

The application of our proposed scheme to compressive sensing of images proceeds as follows. We start with a set of full-length training data, which can be collected at an initialization stage when power resources are not constrained. We then derive a masking pattern using the proposed algorithms at the computational platform (likely away from the sensor), and program the sensor to acquire only the pixels contained in the mask for subsequent captures in order to reduce the power consumption under normal operation. The cost of data acquisition (which in terms of power consumption grows with the number of pixels/data dimensions with the current hardware) is the main motivation for our framework, rather than the cost of computation for training or the cost of manifold learning.

As in most examples where compressive sensing is applicable, the goal here is to trade off simple compression at the sensor (in order to reduce the cost of acquisition) by additional computation that can be incurred outside of the sensor.

This paper is organized as follows. After briefly reviewing the relevant literature in Section II, we propose in Section III both optimization problems and greedy algorithms that select a masking pattern as a subset of the dimensions in the high-dimensional space containing the original dataset, with the general goal being to preserve the structure of the dataset that is relevant during manifold learning. In Section IV, we investigate the proposed algorithms in terms of computational complexity. In Section V, we evaluate the proposed algorithms on six different and diverse image datasets for performance comparison with competing and baseline approaches in masking and feature selection, including articulated objects, handwritten digits, images of human subjects, and the eye gaze tracking images which are representative of the computational eyeglasses application. We offer discussions and some directions for future work in Section VI. Finally, concluding remarks are given in Section VII.

## II. BACKGROUND

### A. Manifold Models and Linear Embeddings

A set of data points $\mathcal{X} = \{x_1, x_2, \ldots, x_n\}$ in a high-dimensional ambient space $\mathbb{R}^d$ that have been generated by an $\ell$-dimensional parameter correspond to a sampling of a *manifold* $\mathcal{M} \subset \mathbb{R}^d$. Given the high-dimensional data set $\mathcal{X}$, we would like to find the parameterization that has generated the manifold. One way to discover this parametrization is to embed the high-dimensional data on the manifold to a low-dimensional space $\mathbb{R}^m$ so that the geometry of the manifold is preserved. *Dimensionality reduction methods* are devised so as to preserve such geometry, which is measured by a neighborhood-preserving criteria that varies depending on the specific algorithm.

A *linear embedding* is defined as a linear mapping $\Phi \in \mathbb{R}^{m \times d}$ that embeds the data in the ambient space $\mathbb{R}^d$ into a low-dimensional space $\mathbb{R}^m$. In many applications, linear embeddings are desirable as dimensionality reduction methods due to their computational efficiency and generalizability. The latter attribute renders linear embeddings easily applicable to unseen test data points. *Principal component analysis* (PCA) is perhaps the most popular scheme for linear dimensionality reduction of high-dimensional data [3], [4]. PCA is defined as the orthogonal projection of the data onto a linear subspace of lower dimension $m$ such that the variance of the projected data is maximized. The projection vectors $\{\phi_i\}_{i=1}^m$ are found by solving the sequential problems [4]

$$\phi_i = \arg \max_{\phi_i : \|\phi_i\|_2 = 1} \sum_{\ell=1}^n \left(\phi_i^T x_\ell - \phi_i^T \bar{x}\right)^2 \quad (1)$$
$$\text{subject to} \quad \phi_i \perp \phi_j \quad \forall \; j < i$$

where $\bar{x} = \frac{1}{n}\sum_{i=1}^{n} x_i$ represents the mean of the data, and $\perp$ designates orthogonality. Note that $\Phi = [\phi_1\ \phi_2\ \ldots\ \phi_m]^T$. Conveniently, the solutions to (1) are the sequence of the dominant eigenvectors of the data covariance matrix [3].

*B. Nonlinear Manifolds and Manifold Learning*

Unfortunately, PCA fails to preserve the geometric structure of a *nonlinear manifold*, i.e., a manifold where the mapping from parameters to data is nonlinear. Particularly, since PCA arbitrarily distorts individual pairwise distances, it can significantly change the local geometry of the manifold. Fortunately, several *nonlinear manifold learning methods* can successfully embed the data into a low-dimensional model while preserving such local geometry in order to simplify the parameter estimation process.

*1) Isomap:* The *Isomap* method aims to preserve the pairwise *geodesic distances* between data points [5]. The geodesic distance is defined as the length of the shortest path between two data points $x_i$ and $x_j$ ($x_i, x_j \in \mathcal{M}$) along the surface of the manifold $\mathcal{M}$ and is denoted by $d_G(x_i, x_j)$. Isomap first finds an approximation to the geodesic distances between each pair of data points by constructing a neighborhood graph in which each point is connected only to its $k$ nearest neighbors; the edge weights are equal to the corresponding pairwise distances. For neighboring pairs of data points, the Euclidean distance provides a good approximation for the geodesic distance, i.e., $d_G(x_i, x_j) \approx \|x_i - x_j\|_2$ for $x_j \in \mathcal{N}_k(x_i)$, where $\mathcal{N}_k(x_i)$ designates the set of $k$ nearest neighbors to the point $x_i \in \mathcal{X}$. For non-neighboring points, the length of the shortest path along the neighborhood graph is used to estimate the geodesic distance. Then, multidimensional scaling (MDS) [27] is applied to the resulting geodesic distance matrix to find a set of low-dimensional points that best match such distances. Note that Isomap is a global method, since the manifold structure is defined by geodesic distances that depend on distances between data points throughout the manifold.

*2) Locally Linear Embedding:* As an alternative, the *locally linear embedding* (LLE) method retains the geometric structure of the manifold as captured by locally linear fits [5]. More precisely, LLE computes coefficients of the best approximation to each data point by a weighted linear combination of its $k$ nearest neighbors. The weights $W = [w_{ij}]$ are found such that the squared Euclidean approximation error is minimized [5]:

$$W = \arg\min_{\bar{W}} \sum_{i=1}^{n} \left\| x_i - \sum_{j:x_j \in \mathcal{N}_k(x_i)} \bar{w}_{ij} x_j \right\|_2^2 \quad (2)$$
$$\text{subject to} \sum_{j:x_j \in \mathcal{N}_k(x_i)} \bar{w}_{ij} = 1, \quad i = 1, \ldots, n.$$

LLE then finds a set of points in an $m$-dimensional space that minimizes the error of the local approximations given by the weights $W$. More precisely, LLE finds the set $\mathcal{Y} = \{y_1, y_2, \ldots, y_n\} \subset \mathbb{R}^m$ that minimizes the squared Euclidean error function [5]

$$\mathcal{Y} = \arg\min_{\{\bar{y}_i\}} \sum_{i=1}^{n} \left\| \bar{y}_i - \sum_{j:x_j \in \mathcal{N}_k(x_i)} w_{ij} \bar{y}_j \right\|_2^2 \quad (3)$$
$$\text{subject to} \sum_{i=1}^{n} \bar{y}_i = 0, \quad \frac{1}{n}\sum_{i=1}^{n} \bar{y}_i \bar{y}_i^T = I,$$

where the first and second constraints are to remove the degrees of freedom due to translation and scaling of the coordinates, in order to obtain a unique solution for the embedding. Note that LLE is considered as a local method, since the manifold structure at each point is determined only by neighboring data points.

*C. Linear Embeddings for Nonlinear Manifolds*

An alternative linear embedding[1] approach to PCA is the method of *random projections*, where the entries of the linear dimensionality reduction matrix are drawn independently following a standard probability distribution such as normal Gaussian or Rademacher. It has been shown that such random projections preserve the relevant pairwise distances between data points with high probability [12]–[16], so that manifold learning algorithms can be applied on the dimensionality reduced data with very small distortion. The drawbacks of random projections are two-fold: ($i$) their theoretical guarantees are asymptotic and probabilistic, and ($ii$) random embeddings are independent of the geometric structure of data, and thus cannot take advantage of training data.

Recently, a near-isometric linear embedding method obtained via convex optimization (referred to as *NuMax*) has been proposed [18], [28]. The key concept in NuMax is to obtain an isometry on the set of pairwise data point differences, dubbed *secants*, after being normalized to lie on the unit sphere [18]:

$$\mathcal{S} = \left\{ \frac{x_i - x_j}{\|x_i - x_j\|_2} : x_i, x_j \in \mathcal{M} \right\}. \quad (4)$$

NuMax relies on a convex optimization problem that finds an embedding $\Phi$ with minimum dimension such that the secants are preserved up to a norm distortion parameter $\delta$. More precisely, the search for a linear embedding is cast as the following rank-minimization problem [18]:

$$P^* = \arg\min \quad \text{rank}(P) \quad (5)$$
$$\text{subject to} \quad |s^T P s - 1| \leq \delta \quad \forall\ s \in \mathcal{S}, P \succeq 0.$$

After $P^*$ is obtained, one can factorize $P^* = \Phi^T \Phi$ in order to obtain the desired low-dimensional embedding $\Phi$. We note that the rank of the solution determines the dimensionality of the embedding, and is controlled by the choice of the distortion parameter $\delta \in [0, 1]$. Note also that $s^T P s = \|\Phi s\|_2^2$; thus, the first constraint essentially upper-bounds the distortion incurred by each secant $s \in \mathcal{S}$.

---

[1]We use the expressions linear dimensionality reduction and linear embedding interchangeably.

The problem (5) is NP-hard, but one may instead solve its nuclear norm relaxation, where the rank of $P$ is replaced by its nuclear norm $\|P\|_*$. Since $P$ is a positive semidefinite symmetric matrix, its nuclear norm amounts to its trace, and thus the optimization in (5) is equivalent to a semidefinite program and can be solved in polynomial time.

*D. Connection with Feature Selection*

The problem of image masking design is reminiscent of feature selection in supervised and unsupervised learning [21], [22]. Previous work on feature selection for unsupervised learning problems (such as manifold learning) is mostly focused on clustering [29]. Spectral feature selection (SPEC) is an unsupervised feature selection method based on spectral graph theory [23]. In SPEC, a pairwise instance similarity metric is used in order to select features that are most consistent with the innate structure of the data. In particular, the radial basis function (RBF) kernel, given by $\exp(-\frac{\|x_i-x_j\|^2}{2\sigma^2})$, is used to measure pairwise similarity between data points. An undirected graph is then constructed with data points as vertices and pairwise similarities as edge weights. Following spectral graph theory, the features are selected so as to preserve the spectrum of the resulting Laplacian matrix. Note that the Laplacian score, proposed earlier in [30], is a special case of SPEC. Similarity preserving feature selection (SPFS) further extends SPEC by overcoming its limitation on handling redundant features [24]. In other words, SPFS considers both similarity preservation and correlation among features in order to avoid choosing redundant features.

## III. MANIFOLD MASKING

In this section, we adopt the criteria used in linear and nonlinear embedding algorithms from Section II to develop algorithms that obtain structure-preserving masking patterns for manifold-modeled data. More precisely, we propose algorithms that attempt to preserve the global and local structure of the manifold, respectively, while reducing the number of dimensions (pixels) of the data points (images). To unify notation, we are seeking a masking index set $\Omega = \{\omega_1, \ldots, \omega_m\}$ of cardinality $m$ that is a subset of the dimensions $[d] := \{1, 2, \ldots, d\}$ of the high-dimensional space containing the original dataset.

*A. Mask Selection with Preservation of Global Structure*

Inspired by the optimization approach of NuMax and the neighborhood-preservation notion of Isomap, we formulate a method for manifold masking that aims at minimizing the distortion incurred by pairwise distances of neighboring data points, which are used in the estimation of global geodesic distances.

Recall that Isomap attempts to preserve the geodesic distances rather than Euclidean distances of data points. Since only the Euclidean distances of neighboring data points match their geodesic counterparts (and the geodesic distance between any two points is found as a function of the geodesic distances between the neighboring points), we are interested in devising a masking operator that preserves the pairwise distances of each data point with its $k$ nearest neighbors. This gives rise to the reduced secant set

$$\mathcal{S}_k = \left\{ \frac{x_i - x_j}{\|x_i - x_j\|_2} : i \in [n], x_j \in \mathcal{N}_k(x_i) \right\} \subseteq \mathcal{S}. \quad (6)$$

To simplify notation, we define the masking linear operator $\Psi : x_i \mapsto \{x_i(j)\}_{j \in \Omega}$ corresponding to the masking index set $\Omega$. We also denote the column vectors $a_i$ with entries $a_i(j) = s_i^2(j)$ for all $j \in [d]$ and for each $i \in [|\mathcal{S}_k|]$. Since the secants are normalized, we have $\sum_{j=1}^{d} a_i(j) = 1$ for all $i \in [|\mathcal{S}_k|]$.

Since a masking operator cannot preserve the norm of the secants, we study the behavior of the masked secant norm under a uniform distribution for the masks $\Omega$. Taking expectation of the secant norms after masking over the random variable $\Omega$ yields

$$\begin{aligned}
\mathbb{E}[\|\Psi s_i\|_2^2] &= \mathbb{E}\left[\sum_{j \in \Omega} a_i(j)\right] \quad (7) \\
&\stackrel{(a)}{=} \sum_{\Omega : |\Omega| = m} \mathbb{P}(\Omega) \left(\sum_{j \in \Omega} a_i(j)\right) \\
&\stackrel{(b)}{=} \sum_{\Omega : |\Omega| = m} \frac{1}{\binom{d}{m}} \sum_{j \in \Omega} a_i(j) \\
&= \frac{1}{\binom{d}{m}} \sum_{\Omega : |\Omega| = m} \sum_{j \in \Omega} a_i(j) \\
&\stackrel{(c)}{=} \frac{1}{\binom{d}{m}} \binom{d}{m} \frac{m}{d} \sum_{j=1}^{d} a_i(j) \\
&\stackrel{(d)}{=} \frac{m}{d},
\end{aligned}$$

where $(a)$ is by the definition of expectation, $(b)$ is due to the masks being equiprobable, $(c)$ is due to the fact that each term $a_i(j)$ appears exactly $\binom{d-1}{m-1} = \binom{d}{m} \frac{m}{d}$ times in the double summation since the number of $m$-subsets of the set $[d]$ that include a particular element is $\binom{d-1}{m-1}$, and $(d)$ is due to the fact that the secants are normalized.

Thus, the norms of the secants $s_i \in \mathcal{S}_k$ are inevitably subject to a compaction factor of $\sqrt{\frac{m}{d}}$ in expectation by the masking operator $\Psi$; this behavior bears out empirically when random masks are used for the datasets considered in Section V. As a result, we will aim to find a masking operator $\Psi$ such that for all $s_i \in \mathcal{S}_k$ we obtain $\|\Psi s_i\|_2^2 \approx \frac{m}{d}$. Note that $\|\Psi s_i\|_2^2 = \sum_{j \in \Omega} s_i^2(j) = \sum_{j=1}^{d} s_i^2(j) z(j) = a_i^T z$, where the indicator vector $z$ is defined by

$$z(j) = \begin{cases} 1 & \text{if } j \in \Omega, \\ 0 & \text{otherwise.} \end{cases} \quad (8)$$

In words, the vector $z \in \{0, 1\}^d$ encodes the membership of the masking index set $\Omega \subseteq [d]$. The average and maximum distortion of the secant norms caused by the masking can be expressed in terms of the vector $z$ and the squared secants

**Algorithm 1** Manifold-Aware Pixel Selection for Global Structure (MAPS-Global)
**Inputs:** normalized squared secants matrix $A$, number of dimensions $m$
**Outputs:** masking index set $\Omega$
**Initialize:** $\Omega \leftarrow \{\}$
**for** $i = 1 \to m$ **do**
    $\bar{A}_\Omega \leftarrow A_\Omega \cdot 1_{|\Omega|}$          {compute current masked secant squared norms}
    $\omega_i \leftarrow \arg\min_{\omega \in \Omega^c} \|A_\omega + \bar{A}_\Omega - \frac{i}{d} 1_{|\mathcal{S}_k|}\|_p$     {minimize aggregate difference with $\mathbb{E}[\|\Psi s_i\|_2^2]$}
    $\Omega \leftarrow \Omega \cup \{\omega_i\}$          {add selected dimension to the masking index set}
**end for**

matrix $A := [a_1\ a_2\ \ldots\ a_{|\mathcal{S}_k|}]^T$ as follows:

$$\sum_{s_i \in \mathcal{S}_k} \left| \|\Psi s_i\|_2^2 - \frac{m}{d} \right| = \left\| Az - \frac{m}{d} 1_{|\mathcal{S}_k|} \right\|_1, \quad (9)$$

$$\max_{s_i \in \mathcal{S}_k} \left| \|\Psi s_i\|_2^2 - \frac{m}{d} \right| = \left\| Az - \frac{m}{d} 1_{|\mathcal{S}_k|} \right\|_\infty, \quad (10)$$

respectively, where $1_{|\mathcal{S}_k|}$ denotes the $|\mathcal{S}_k|$-dimensional all-ones column vector. Thus, in order to leverage the metrics (9-10) during selection, we propose to find the optimal masking pattern by casting the following integer program:

$$z^* = \arg\min_z \quad \left\| Az - \frac{m}{d} 1_{|\mathcal{S}_k|} \right\|_p \quad (11)$$
$$\text{subject to} \quad 1_d^T z = m, z \in \{0,1\}^d,$$

where $p = 1$ and $p = \infty$ correspond to optimizing the average and maximum secant norm distortion caused by the masking, respectively.[2] The equality constraint dictates that only $m$ dimensions are to be retained in the masking process.

The integer program (11) is computationally intractable even for moderate-size datasets [31]. We note that the non-integer relaxation of (11) results in the trivial solution $z^* = \frac{m}{d} 1_d$. Note also that the matrix $A$ depends on the dataset used; thus in general it does not hold necessary properties for relaxations of integer programs to be successful (e.g. being totally unimodular, having binary entries, etc.). We also attempted a Lagrangian non-integer relaxation of the form

$$z^* = \arg\min_z \left\| Az - \frac{m}{d} 1_{|\mathcal{S}_k|} \right\|_p + \lambda \|z\|_1, \quad (12)$$

where again $p = 1$ or $p = \infty$. Note that since this is a non-integer relaxation, we consider the sparsity pattern of the solution to obtain a mask. We observed that (a) the performance is worse than that obtained by the IP, and (b) it is difficult to obtain the value of the Lagrangian multiplier needed for a particular mask size.

We propose a heuristic greedy algorithm that can find an approximate solution for (11) in a drastically reduced time. The greedy approach in Algorithm 1, which we refer to as Manifold-Aware Pixel Selection for Global structure (MAPS-Global), gives an approximate solution for the $\ell_p$-norm minimization in (11). The algorithm iteratively selects elements of the masking index set $\Omega$ as a function of the squared secants matrix $A$. We initialize $\Omega$ as the empty set and denote $\Omega^c = [d] \setminus \Omega$. At iteration $i$ of the algorithm, we find a new dimension that, when added to the existing dimensions in $\Omega$, causes the squared norm of the masked secant to match the expected value of $\frac{i}{d}$ as closely as possible. More precisely, at step $i$ of the algorithm, we find the column of $A$ indexed by $\omega \in \Omega^c$ (which is indicated by $A_\omega$), whose addition with the sum of previously chosen columns $\bar{A}_\Omega = \sum_{\omega \in \Omega} A_\omega$ has minimum distance (in $\ell_p$-norm) to $\frac{i}{d} 1_{|\mathcal{S}_k|}$. Note that $\bar{A}_\Omega = Az$, where $z$ again denotes the indicator vector for the masking index set $\Omega \subseteq [d]$; thus, the metric guiding the greedy selection matches the objective function of the integer program (11).

*B. Mask Selection with Preservation of Local Structure*

Next, we propose a greedy algorithm for selection of a masking pattern that attempts to preserve the local structure of the manifold. The idea of this algorithm is to preserve the weights $w_{ij}$ obtained from the optimization in (2). Preserving these weights would in turn maintain the embedding $\mathcal{Y}$ found from (3) through the image masking process.

The rationale behind the proposed algorithm is as follows. The weights $w_{ij}$ for $j \in \mathcal{N}_k(x_i)$ are preserved if both the lengths of the secants involving $x_i$ (up to a scaling factor) and the angles between these secants are preserved. Geometrically, this can be achieved if the distances between all the points in the set $\mathcal{C}_{k+1}(x_i) := \mathcal{N}_k(x_i) \cup \{x_i\}$ are preserved up to a scaling factor. For this purpose, we define the *secant clique* for $x_i$ as

$$\mathcal{S}_{k+1}(x_i) := \{x_{j_1} - x_{j_2} : x_{j_1}, x_{j_2} \in \mathcal{C}_{k+1}(x_i)\}; \quad (13)$$

our goal for the mask selection is to preserve the norms of these secants up to a scaling factor. This requirement can be captured by a normalized inner product commonly referred to as *cosine similarity measure*, defined as $\text{sim}(\alpha, \beta) := \frac{\langle \alpha, \beta \rangle}{\|\alpha\|_2 \|\beta\|_2}$. To implement our method, we define a 3−dimensional array $B$ of size $c \times d \times n$, where $c = \binom{k+1}{2}$ denotes the number of elements in each secant clique $\mathcal{S}_{k+1}(x_i)$. The array has entries $B(\ell, j, i) = s_\ell^i(j)^2$, where $s_\ell^i$ denotes the $\ell^{\text{th}}$ secant contained in $\mathcal{S}_{k+1}(x_i)$. In words, every 2-D slice of $B$, denoted by $B_i := B(:,:,i)$ corresponds to the squared secants matrix for the secant clique $\mathcal{S}_{k+1}(x_i)$, and the $\ell^{\text{th}}$ row of $B_i$ corresponds to the $\ell^{\text{th}}$ secant in $S_{k+1}(x_i)$.

We now define our mask metric used to preserve the local manifold structure. The vector $\alpha = B_i z$, where $z$ is

---
[2] Note that we have also tried $p = 2$ numerically, but the masks obtained do not preserve the desired manifold structure. Also, note that we tried considering a scaling factor $\gamma$ as an optimization parameter in (11) in place of the constant $\frac{m}{d}$, but the latter performed best.

**Algorithm 2** Manifold-Aware Pixel Selection for Local Structure (MAPS-Local)

**Inputs:** neighborhood clique secant array $B$, masking size $m$
**Outputs:** masking index set $\Omega$
**Initialize:** $\Omega \leftarrow \{\}$
$\alpha \leftarrow \sum_{j \in [d]} B(:,j,:)$ {Compute matrix of squared secant norms.}
**for** $i = 1 \to m$ **do**
  $\theta \leftarrow \sum_{j \in \Omega} B(:,j,:)$ {Compute matrix of squared masked secant norms for current masking set $\Omega$.}
  **for** $j \in \Omega^C$ **do**
    $\beta \leftarrow \theta + B(:,j,:)$ {Update squared masked secant norms when $\{j\}$ is added to mask $\Omega$.}
    $\lambda(j) \leftarrow \sum_{t \in [n]} \frac{\langle \alpha(:,t), \beta(:,t) \rangle}{||\alpha(:,t)||_2 ||\beta(:,t)||_2}$ {Compute cosine similarity measure for updated mask.}
  **end for**
  $\omega \leftarrow \arg\max_{j \in \Omega^C} \lambda(j)$ {Find new mask element that maximizes cosine similarity.}
  $\Omega \leftarrow \Omega \cup \{\omega\}$ {Add selected dimension to the masking index set.}
**end for**

---

the mask indicator vector from (8), contains the squared norms of the masked secants from $S_{k+1}(x_i)$ as its entries. Similarly, the vector $\beta = B_i 1_d$ will contain the squared norms of the full secants in the same set. Maximizing the cosine similarity $\text{sim}(\alpha, \beta)$ promotes these two vectors being a scaled version of one another, i.e., the norms of the masked secants approximately being equal to a scaling of the full secant norms. Note that since we are aiming for preservation of the local manifold structure, the value of this scaling can vary over data points without incurring distortion of the manifold structure. In order to incorporate the cosine similarity measure for all data points, we maximize the sum of the aforementioned similarities for all data points as follows:

$$\hat{z} = \arg\max_z \sum_{i=1}^n \frac{\langle B_i z, B_i 1_d \rangle}{||B_i z||_2 ||B_i 1_d||_2} \quad (14)$$
$$\text{subject to} \quad 1_d^T z = m, z \in \{0,1\}^d.$$

Finding an optimal solution for $z$ from (14) has a combinatorial (exponential) time complexity. An approximation can be obtained by greedily selecting the masking elements that maximize the value of the mask metric, one at a time. The proposed algorithm, which we call Manifold-Aware Pixel Selection for Local structure (MAPS-Local), is given in Algorithm 2.

## IV. Computational Complexity of Masking

In this section, we compare different masking methods in terms of computational complexity. Note that in the setup that we are considering, the computational complexity of mask selection is not as important as the performance of the chosen mask. As mentioned in Section I, this is because the selection of masks are essentially done in a training stage where computational resources are not constrained.

The computational complexity of MAPS-Global is $\mathcal{O}(mdkn)$. To see this, note that in each of the $m$ iterations the search for $\omega \in \Omega^c$ considers at most $d$ elements, and the number of arithmetic operations in computing the $\ell_p$-norm term is $\mathcal{O}(|S_k|)$. Thus, we have

$$T_{\text{MAPS-Global}}(n,m,k,d) = \mathcal{O}(md|S_k|) = \mathcal{O}(mdkn), \quad (15)$$

where the last equality uses the fact that $|S_k| \leq kn$.

The computational complexity of MAPS-Local is $\mathcal{O}(mk^2 nd)$. To see this, note that the complexity of computing each of the matrices $\alpha$, $\theta$, and $\beta$ is proportional to the number of elements of the array $B$ involved in the summation; thus the aforementioned complexities are $\mathcal{O}(cdn)$, $\mathcal{O}(cmn)$, and $\mathcal{O}(cn)$, respectively. In addition, the computation of the cosine similarity vector $\lambda$ can be done in $\mathcal{O}(cn)$ time. As a result, the complexity of MAPS-Local is given by

$$T_{\text{MAPS-Local}}(n,m,k,d) = \mathcal{O}(cdn) + \mathcal{O}(m)\Big(\mathcal{O}(cmn) \quad (16)$$
$$+ \mathcal{O}(d)\big(\mathcal{O}(cn) + \mathcal{O}(cn)\big)\Big)$$
$$= \mathcal{O}(cdn) + \mathcal{O}(cm^2 n)$$
$$+ \mathcal{O}(mcnd)$$
$$\stackrel{(a)}{=} \mathcal{O}(mcnd)$$
$$\stackrel{(b)}{=} \mathcal{O}(mk^2 nd),$$

where in $(a)$ we exploit the fact that $m < d$ and $(b)$ is due to $c = \mathcal{O}(k^2)$.

The computational complexity of random masking (random subset selection) is $\mathcal{O}(d)$ via Algorithm R [32]. The computational complexities of feature selection algorithms SPEC [23] and SPFS [24] are $\mathcal{O}((n+d)n^2)$ and $\mathcal{O}(dnm^2)$, respectively. Note that we used the SPFS-LAR version of SPFS which performs best among other SPFS versions and does not require parameter tuning. For Sparse PCA we used the implementation given by [33], which has the computational complexity of $\mathcal{O}((d^2 + nd)t)$, where $t$ is the number of iterations required for the algorithm to converge.

Random masking is by far the fastest algorithm, but it comes at the price of its poor performance for the majority of the datasets and experimental settings. Note that the complexity for all the other algorithms includes a factor of $\mathcal{O}(nd)$ (as lower bound); the remaining terms for different algorithms depend on the experimental settings, but are roughly comparable. For instance, at lower mask sizes MAPS-Global is one of the faster algorithms along with Sparse PCA, considering that the number of iterations for the algorithm under consideration for SPCA is in the order of a few hundreds. Additionally, if the number of dimensions $d$ and/or the number of data points $n$ are sufficiently high, the proposed MAPS-Global has an advantage over the rest of the algorithms since the remaining factor is a function of neither $n$ nor $d$.

## V. NUMERICAL EXPERIMENTS

In this section, we present a set of experimental results that compare the performance of the proposed algorithms to those in the existing linear embedding and feature selection literatures, in terms of preservation of the low-dimensional structure of several nonlinear manifolds.[3]

We once again remark that the goal of the masking schemes proposed here is to reduce the number of data dimensions (in order to reduce data acquisition costs) while preserving the manifold structure. Thus, if we apply a manifold learning algorithm (e.g., Isomap) on the masked data, the resulting embedding is ideally as close as possible to that obtained from full data. In addition, having obtained the embedding of the masked images from a manifold, we would like to evaluate how well the embedding can be extended to new masked images — a setup known in the literature as out-of-sample extension [34]. Thus, our comparison with standard dimensionality reduction schemes aims to show whether a performance gap exists if manifold learning schemes are applied to the masked images versus the original (full) images.

### A. Experimental Setup and Comparison Methods

We evaluate the methods described in Section III. In addition, we consider *random masking*, PCoA (described in the sequel), Sparse PCA [35], [36], and two unsupervised feature selection methods, SPEC [23] and SPFS [24]. In random masking, we pick an $m$-subset of the $d$ data dimensions uniformly at random. Sparse PCA (SPCA) is a variation of PCA in which sparsity is enforced in the principal components. Note that since the support of the principal components is not required to be the same, we focus on the support of the first principal component so that we can translate Sparse PCA into a masking scheme. For Sparse PCA we use the implementation given in [33]. Note also that we use the SPFS-LAR version of SPFS, which is favored by the authors of SPFS, since it does not require extra parameter tuning (other than the parameter $\sigma$ of the RBF kernel function). In our experiments, we perform a grid search over $\{1, 2, \ldots, 10\}$ in order to find the value of the parameter $\sigma$ that works best.

*Principal coordinate analysis* (PCoA) is a natural adaptation of PCA to mask design. The main idea of PCoA is to find the $m$ canonical basis vectors (rather than arbitrary orthogonal vectors in PCA) that span the canonical subspace which captures the highest variance of the data through projection. Substituting $\phi_i$ with canonical basis elements $e_i$ in (1) yields

$$\omega_i = \arg \max_{i \in [d]} \sum_{\ell=1}^{n} (x_\ell(i) - \bar{x}(i))^2 \quad (17)$$
$$\text{subject to} \quad \omega_i \neq \omega_j \quad \forall \ j < i,$$

and so the masking pattern $\Omega$ is found by solving (17) sequentially for $i = 1, \ldots, m$. In practice, this masking pattern can be obtained greedily by selecting the indices of the $m$ dimensions with the highest variances across the dataset. Note that the computational complexity of PCoA is given by $\mathcal{O}(dn + d \log(d))$.

For our experiments, we use five standard manifold modeling datasets — the *MNIST* dataset [37], the *Heads* dataset [38],[4] the *Faces* dataset [7], the *Statue* dataset [39]–[41], and the *Hands* dataset [5] — as well as one custom eye-tracking dataset from a computational eyeglass prototype as detailed in Table I. For the MNIST dataset, we focus on the subset corresponding to the handwritten digit 2's. The *Eyeglasses* dataset corresponds to captures from a prototype implementation of computational eyeglasses that use the imaging sensor array of [20].

The algorithms listed above are tested for maskings of size $m = 50, 100, 150, 200, 250, 300$. For PCA,[5] $m$ provides the dimensionality of the embedding (i.e., we apply PCA to full-length data to obtain an embedding of dimensionality $m$). Note that since PCA employs all $d$ dimensions of the original data, it has an intrinsic performance advantage over the masking algorithms. The performance of random masking is averaged over 100 independent draws in each case.

The combinatorial nature of the integer program (11) renders it significantly expensive in computation, even for the small dimensions for the data shown in Table I (not converging even after 24 hours in our experiments). In contrast, the remaining masking algorithms each take only up to 20 seconds (for $m = 300$) to complete using the same computing platform. In [1], MAPS-Global has been shown to be a good approximation of the integer program. Hence, here we only consider MAPS-Global in our experiments.

Figure 1 indicates the masking patterns associated with different masking methods for all the datasets for a mask size of $m = 100$ pixels; the active pixels (i.e. the pixels that are preserved by the mask) are marked in white. As shown in this figure, MAPS-Global and MAPS-Local do not select the pixels with the highest variance, in contrast to PCoA. The pixel masks selected by the MAPS algorithms suggest that pixels with highest variations are not necessarily more informative of the underlying manifold structure.

We note in passing that in certain LLE experiments we obtained data covariance matrices that are singular or nearly singular (often due to masking). In such cases, the covariance matrix can be conditioned by adding a small multiple of the identity matrix [42], [43].

---

[3]MATLAB code for generation of the results of this section is available at http://www.ecs.umass.edu/~mduarte/Software.html.

[4]This dataset is originally termed as the Faces dataset. However, in order to avoid confusion with the Faces dataset of [7], we rename it to the Heads dataset.

[5]We excluded NuMax from consideration since its performance for linear embeddings of dimensions between $d$ and $m$ (which is moderately large here) is similar to that of PCA for our datasets.

TABLE I
SUMMARY OF EXPERIMENTAL DATASETS

| Dataset | **Eyeglasses** | **MNIST** | **Statue** | **Heads** | **Faces** | **Hands** |
|---|---|---|---|---|---|---|
| Number of images $n$ | 929 | 1000 | 960 | 698 | 1965 | 1000 |
| Embedding dim. $\ell$ | 2 | 5 | 3 | 3 | 3 | 4 |
| Image dim. $d$ | $40 \times 40$ | $28 \times 28$ | $51 \times 34$ | $32 \times 32$ | $28 \times 20$ | $64 \times 64$ |
| Neighborhood size for Isomap $k$ | 12 | 10 | 12 | 10 | 9 | 8 |
| Neighborhood size for LLE $k$ | 12 | 10 | 12 | 7 | 10 | 10 |

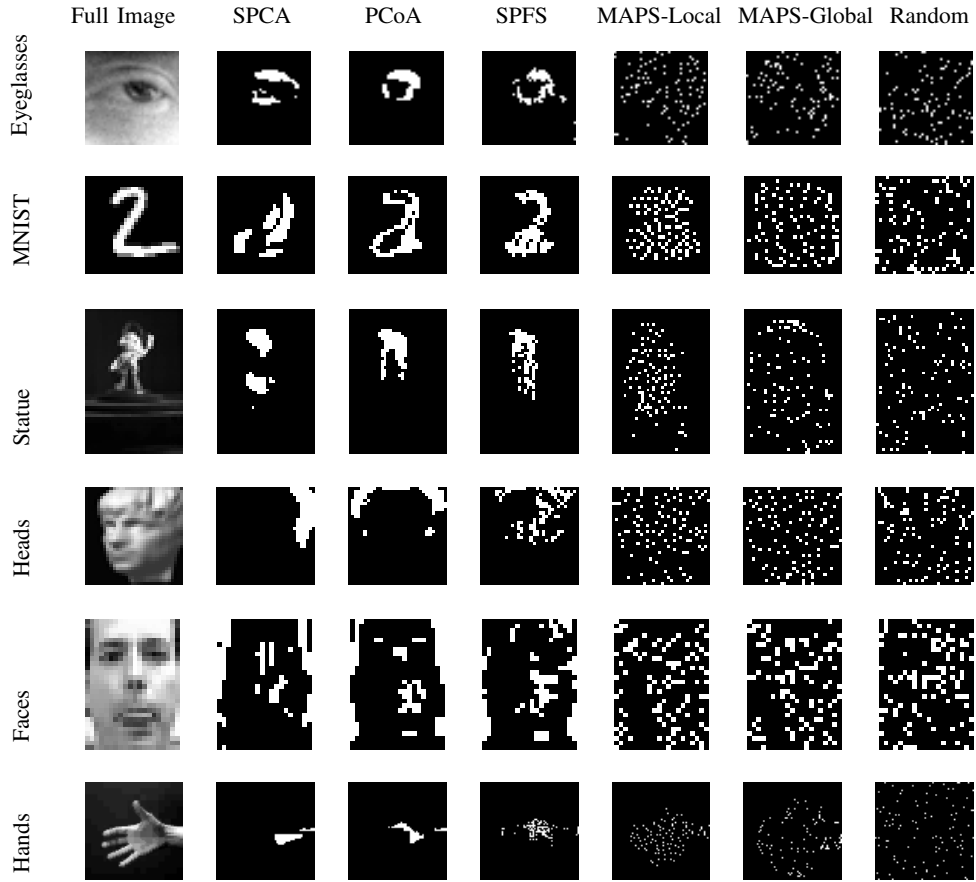

Fig. 1. Masks obtained for each dataset with masking size of $m = 100$ via different masking schemes.

For each selection of masking algorithm and size, we apply the manifold learning algorithm (either Isomap or LLE) directly on the masked images. We then check the performance of the manifold embedding obtained from the masked datasets to that of the manifold embedding from the full dataset using different performance metrics.

Table II shows sample running times (in seconds) for different masking algorithms when applied to different datasets. The results agree for the most part with what would be expected from the provided computational complexities. As can be observed from this table, the baseline algorithms (random masking and PCoA) are faster than other algorithms. As expected, MAPS-Local has a higher computational complexity than the other options. The computational complexity of the remaining algorithms are comparable, with MAPS-Global and Sparse PCA being faster than competing feature selection algorithms (i.e., SPEC and SPFS) for most of the datasets.

### B. Masking with Preservation of Global Structure

In order to evaluate the preservation of the global manifold structure, we use the following two criteria to evaluate the performance of masking or embedding methods. First, we use *residual variance* as a global metric to measure how well the Euclidean distances in the embedded space match the geodesic distances in the ambient space [5]. For each dataset, we pick the embedding dimensionality $\ell$ to be the value after which the residual variance ceases to decrease substantially with added dimensions. Note that the obtained values of $\ell$ agree with the intuitive number of degrees of freedom for the Heads dataset (two rotation angles – pitch and yaw – for orientation, plus an illumination variable), the Eyeglasses dataset (2-D gaze locations), and the Statue dataset (2-D rotation plus camera position). Second, we use the percentage of preserved nearest neighbors [18]. More precisely, for a given neighborhood of size $k$, we obtain the percentage of the $k$-nearest neighbors in the full $d$-dimensional data that are among the $k$-nearest neighbors when the masked image manifold is embedded.

TABLE II
COMPARISON OF COMPUTATION TIMES (IN SECONDS) FOR DIFFERENT ALGORITHMS AT MASKING SIZE $m = 50$

| Dataset | Eyeglasses | MNIST | Statue | Heads | Faces | Hands |
|---|---|---|---|---|---|---|
| MAPS-Global | 5.32 | 3.06 | 4.03 | 2.37 | 3.65 | 18.24 |
| MAPS-Local | 72.62 | 53.02 | 76.31 | 21.73 | 46.53 | 177.10 |
| Random | 0.00027 | 0.00020 | 0.00022 | 0.00021 | 0.00021 | 0.00037 |
| PCoA | 0.021 | 0.019 | 0.020 | 0.0081 | 0.026 | 0.083 |
| Sparse PCA | 3.81 | 2.89 | 3.49 | 3.68 | 1.48 | 29.46 |
| SPFS | 6.31 | 4.38 | 7.31 | 4.09 | 4.09 | 19.37 |
| SPEC | 5.56 | 4.84 | 6.81 | 1.33 | 14.16 | 17.14 |

In Figures 2 and 3, we display the residual variance and neighborhood preservation results of different masking and embedding methods, respectively, when Isomap is used as the manifold learning algorithm. More precisely, in Figure 2, for each given dataset, we vary the masking size $m$ from 50 to 300 and measure the residual variance of the manifold learned from the masked data (or data with dimensionality reduced to m). Note that for each point in the plot, a new manifold is learned from a subset of image pixels of size $m$. Also, note that each mask of size $m_1$ is a superset of all the masks of size $m_2 < m_1$, i.e., the masks are incremental or nested inside one another. MAPS-Global is shown only for the choice $p = 1$, as setting $p = \infty$ yields similar results. We observe that the performance of MAPS-Global and MAPS-Local are significantly and consistently better than those of random sampling, PCoA, and Sparse PCA. PCoA fails to identify the best dimensions to preserve from the original data. This failure is particularly evident for the Heads dataset, where the distribution of the image energy across the pixels is most uniform. SPCA has an erratic behavior across datasets; it is performing well for some of the datasets and for moderately large values of $m$, but poorly for other datasets and for lower values of $m$. Note that, as expected, SPFS always outperforms SPEC, but is outperformed by our MAPS algorithms. Additionally, we have dropped the curve related to SPEC for datasets for which SPEC was performing poorly. The values of the parameter $\sigma$ used for SPFS is $[6, 2, 4, 4, 5]$ for eyeglasses, MNIST, Heads, Faces, and Statue datasets, respectively. Interestingly, random masking outperformed all the methods other than the proposed MAPS algorithms for Heads and Faces datasets. This can be attributed to the activity being more spread out over the pixels for the latter datasets.

For small values of $m$, PCA can significantly outperform the masking algorithms of Section III, which is to be expected since the former employs all $d$ dimensions of the original data. More surprisingly, we see that for sufficiently large values of $m$ the performance of the MAPS algorithms approaches or matches that of PCA, even though the embedding feasible set for masking methods is significantly reduced. The results are consistent across the datasets used in the experiments.

As an example of both performance and savings obtained by masking, note that for the eyeglasses dataset, the performance of the masked manifolds via the proposed MAPS algorithms (when preserving the global manifold structure) is essentially the same as that of full data (with $d = 1600$ pixels) at masking size $m = 200$ pixels. As a result, from [17], we are reducing the power consumption of the camera from 30 mW to 8 mW (operating at 4 frames per second), when sensing each eye image, thereby reducing the power consumption of the camera significantly.

Finally, we compare the performance of different masking schemes at preserving the 2-D manifolds learned (via Isomap) from the Eyeglasses dataset, containing pictures of an eye pointed in different directions, and the Heads dataset, in which a 3-D model of a head is subject to rotations in pitch and yaw. As shown in Figure 4, the 2-D manifold learned from images masked using MAPS-Global with $m = 50$ pixels resembles the 2-D manifold learned from full images. We have also verified that when the size of the mask is increased to $m = 200$, the 2-D manifold learned from the masked images is essentially visually identical to that learned from the full data. On the other hand, the masks chosen using random masking, SPCA, and PCoA warp the structure of the manifold learned from the masked data, which creates shortcuts between the left and right hand sides of the manifold.

### C. Masking with Preservation of Local Structure

In order to evaluate the preservation of the local manifold structure, we consider the following *embedding error*. Suppose the pairs $(\mathcal{X}, \mathcal{Y})$ and $(\mathcal{X}', \mathcal{Y}')$ designate the ambient and embedded set of vectors for full and masked data, respectively. Having found the weights $w_{i,j}$ from the full data via (2), we define the embedding error for the masked data in the following way:

$$e = \sum_{i=1}^{n} \left\| y'_i - \sum_{j: x_j \in \mathcal{N}_k(x_i)} w_{ij} y'_j \right\|_2^2. \quad (18)$$

The rationale behind this definition of the embedding error is that, ideally, the embedded vectors $y'_i$ obtained from masked images should provide a good linear fitting using the neighborhood approximation weights obtained from the original (full) images. In other words, (18) finds the amount of deviation of $\mathcal{Y}'$ from $\mathcal{Y}$, which minimizes the value of this score, cf. (3).

Since local manifold learning algorithms (such as LLE) do not preserve the global structure of the manifold, there is no guarantee for preservation of nearest neighbors beyond $k$ in general. This was observed in our experiments by the non-monotonicity of neighborhood preservation as a function of the masking/embedding size. Thus for LLE we do not include plots for percentage of preserved nearest neighbors.

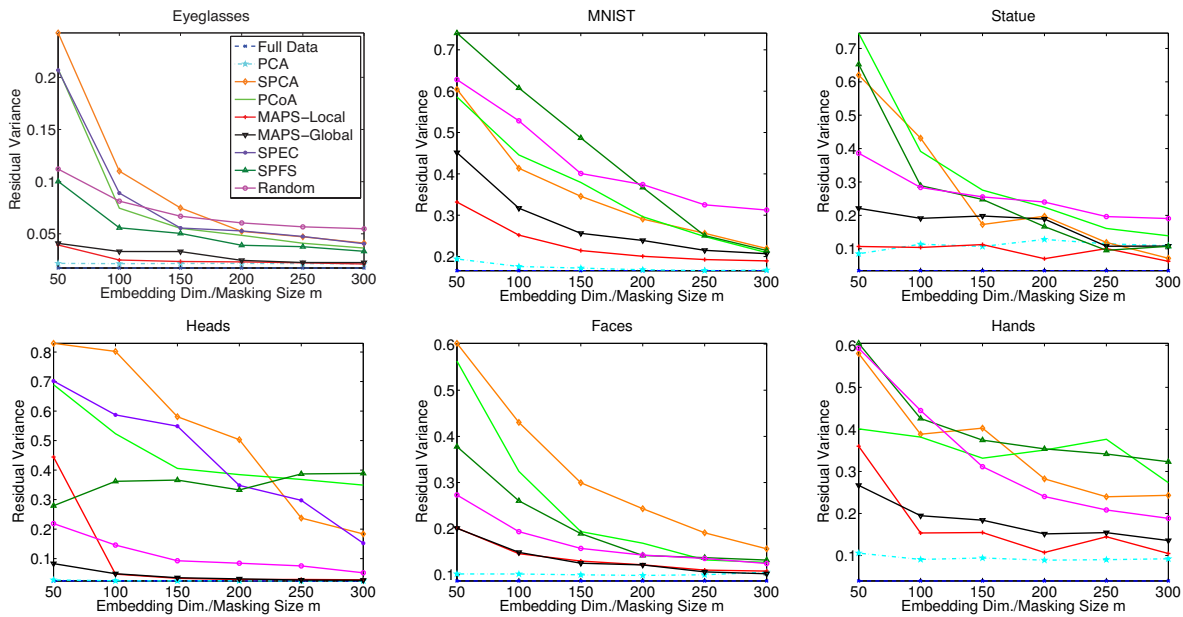

Fig. 2. Performance comparison in terms of residual variance for linear embeddings (dashed lines) and masking algorithms (solid lines) with respect to original full-length data, when Isomap is used as the manifold learning algorithm. Residual variance as a function of $m$ is used.

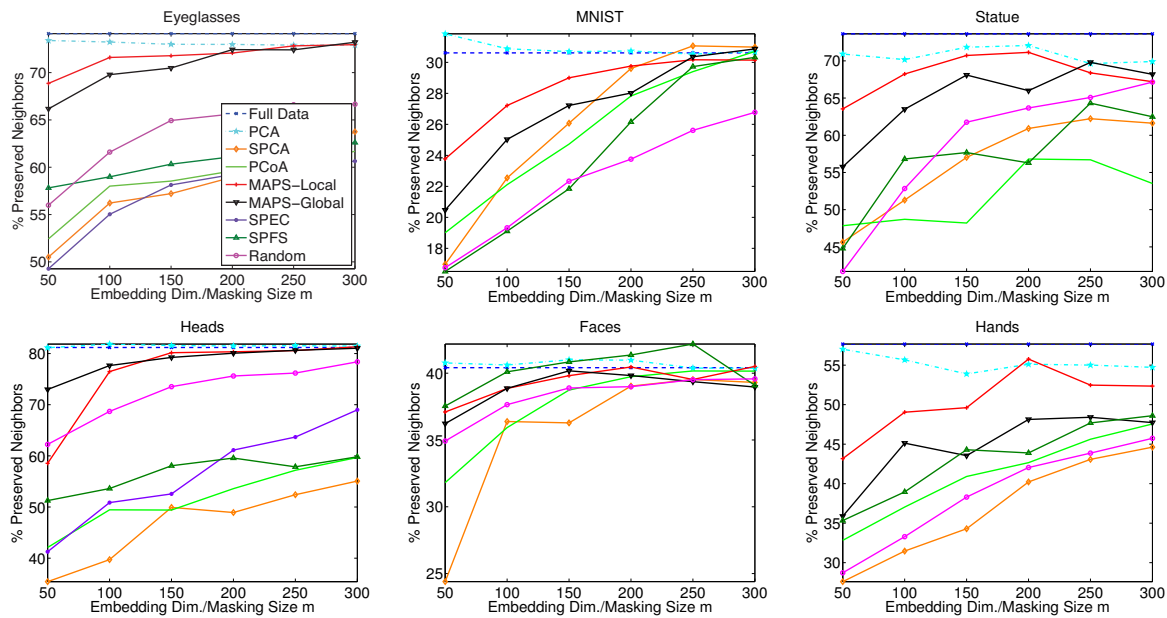

Fig. 3. Performance comparison for linear embeddings (dashed lines) and masking algorithms (solid lines) with respect to original full-length data, when Isomap is used as the manifold learning algorithm. Percentage of preserved nearest neighbors for 20 neighbors is used.

Figure 5 shows the embedding error plots over different datasets for the case that LLE is used as the manifold learning algorithm. Here we can see that the MAPS-Local algorithm outperforms all the other masking algorithms across all the datasets consistently. Note that for the plots of Heads and Faces datasets, we have dropped the SPCA curves due to its poor performance and change in the scaling of the plots as a result. The values of the parameter $\sigma$ used for SPFS is $[5, 4, 4, 4, 9]$ for eyeglasses, MNIST, Heads, Faces, and Statue datasets, respectively.

Next, we compare the performance of different masking schemes at preserving the 2-D manifolds learned (via LLE) from the Eyeglasses dataset. We compare the performance of MAPS-Local in preserving the 2-D manifold from Eyeglasses dataset with that of random masking, SPCA, and PCoA at masking size of $m = 100$. As can be observed from Figure 6, the 2-D manifold learned from images masked via MAPS-Local resembles that learned from the full images more closely than PCoA. Note that the manifold learned from random masking is warped and does not preserve the distances among data points faithfully. Also note that the LLE embedding for the Heads dataset does not provide a clear visualization of the controlled parameters.

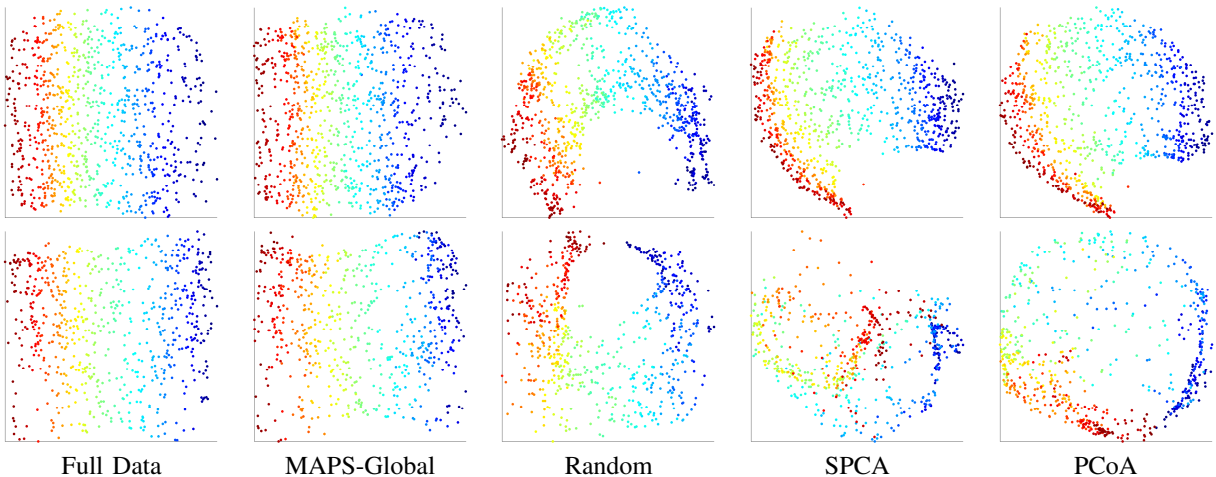

Fig. 4. Performance comparison of two-dimensional projections of eyeglasses (top row) and heads dataset (bottom row) masked with $m = 50$ via different methods after Isomap manifold learning.

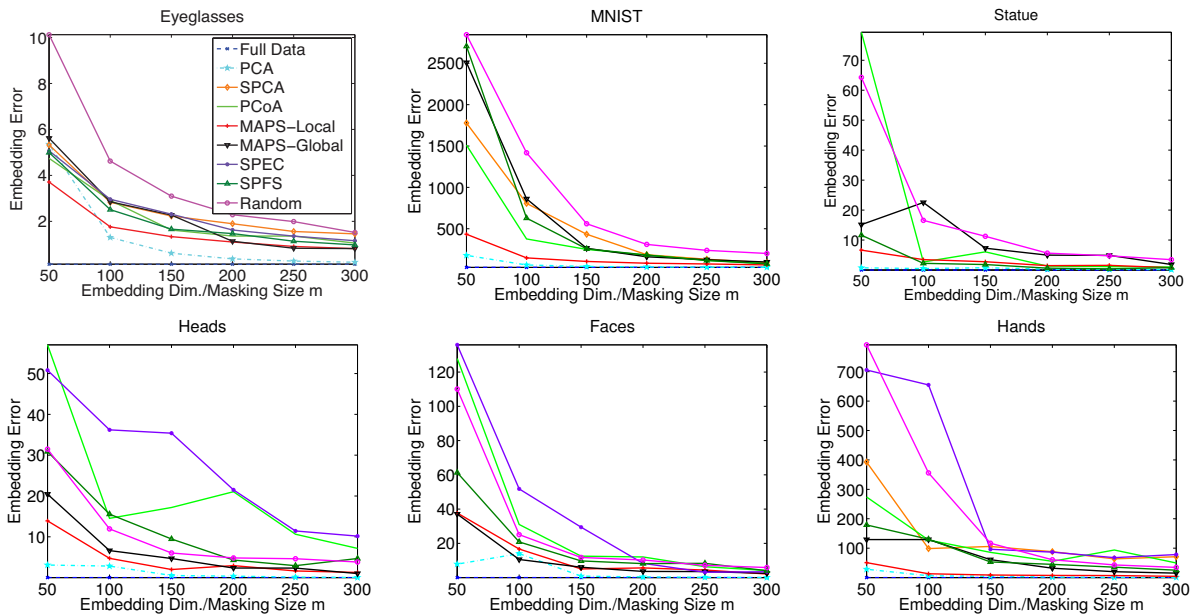

Fig. 5. Performance comparison for linear embeddings (dashed lines) and masking algorithms (solid lines) with respect to original full-length data when LLE is used as the manifold learning algorithm. Embedding error as a function of $m$.

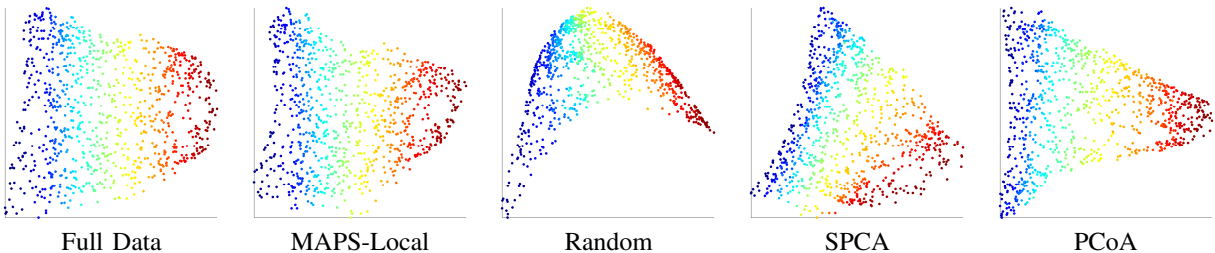

Fig. 6. Performance comparison of two-dimensional projections of eyeglasses dataset masked with $m = 100$ via different methods after LLE manifold learning.

### D. Masking and Out-of-Sample Extension

Next, we consider the effect of masking on out-of-sample extension (OoSE) for manifold learning algorithms. OoSE generalizes the result of the nonlinear manifold embedding for new data points. For LLE OoSE, we use the procedure derived in [42], [44]. For Isomap OoSE, we use the procedure suggested in [25], [44].

The experiments in this section pursue the following general framework. First, we apply the masks designed in the previous section on the dataset. Next, we perform OoSE in a leave-one-out fashion on the masked dataset excluding the selected data point. Then, we compare the embedding for the new data point to its counterpart obtained from embedding of all the data points from full data (including the point which is left out as an "out of sample").

Note that for Isomap we cannot directly compare these two points [45], as embeddings learned from different samplings of the manifold are often subject to translation, rotation, and scaling. These variations must be addressed via manifold alignment before the embedded datasets are compared. We find the optimal alignment of the original manifold and the OoSE manifold via *Procrustes analysis* [46], [47] and apply the resulting translational, rotational, and scaling components on the OoSE manifold. Finally, we measure the OoSE error as the $\ell_2$ distance between the two manifolds for the embedded test point, averaged across all test points.

Due to the local nature of LLE, the embedding obtained via OoSE remains unchanged from the original for most of the data points. Hence, it is logical to only consider the embedding error for the points that are affected by OoSE. Let $x_{i_0}$ indicate the out-of-sample point, and define the set $\mathcal{N}'_k(x_{i_0})$ of points affected by OoSE on point $x_{i_0}$ as

$$\mathcal{N}'_k(x_{i_0}) = \{x_i \in \mathcal{X} : x_{i_0} \in \mathcal{N}_k(x_i) \text{ or } x_i = x_{i_0}\}, \quad (19)$$

i.e., the set of all the points that have $x_{i_0}$ as their neighbors plus $x_{i_0}$ itself. Denote the set of indices for points contained in $\mathcal{N}'_k(x_{i_o})$ as $\mathcal{I}(i_0) = \{i : x_i \in \mathcal{N}'_k(x_{i_o})\}$. We then define a version of the metric (18) that accounts only for local linear fits of the affected by OoSE as

$$e_{\text{OoSE}} = \frac{1}{n} \sum_{i_o=1}^{n} \sum_{i \in \mathcal{I}_o} \left\| y'_i - \sum_{j : x_j \in \mathcal{N}_k(x_i)} w_{ij} y'_j \right\|_2^2, \quad (20)$$

which we term as average OoSE embedding error.

Figures 7 and 8 show the performance of OoSE from masked images for Isomap and LLE as the manifold learning algorithm, respectively. In each case, due to the high computational complexity of the leave-one-out experiment in this setting, we only compare the performance of the respective MAPS algorithm with that of random masking. As can be observed from the figures, for both Isomap and LLE OoSE, the respective MAPS algorithms consistently outperform random masking for all datasets.

### E. Application of Masking in Eye Gaze Estimation

Finally, we consider an application of manifold models in our motivating computational eyeglasses platform. More precisely, we focus on the Eyeglasses dataset, illustrated in Figure 9, which is collected for the purpose of training an estimation algorithm for eye gaze position in a 2-D image plane. The dataset corresponds to a collection of image captures of an eye from a camera mounted on an eyeglass frame as the subject focuses their gaze into a dense grid of known positions (size $31 \times 30$, covering a $600 \times 600$ pixel screen projection) that is used as ground truth.

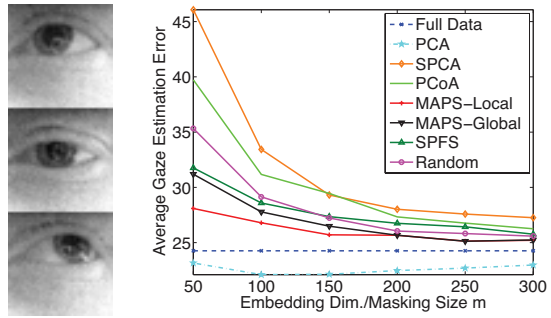

Fig. 9. Left: Example images from *Eyeglasses* dataset. Right: Performance of eye gaze estimation using an appearance-based method from embedded and masked images as a function of $m$.

Most of the literature on eye gaze estimation has focused on *feature-based* approaches, where explicit geometric features such as the contours and corners of the pupil, limbus and iris, are used to extract features of the eye [48], [49]. Unfortunately, such methods require all the pixels of the eye image and are therefore not compatible with image masking. Alternatively, an *appearance-based* method that adopts the nonlinear manifold models at the center of this paper has been proposed in [26]. The idea behind this method is to find a nonlinear manifold embedding of the original dataset $\mathcal{X}$ and extend it to the 2-D parameter space samples given by the eye gaze ground truth. The proposed method employs the weights obtained by LLE, when applied to the training image dataset together with a testing image $\mathcal{X} \cup \{x_t\}$, and applies these weights in the parameter space to estimate the parameters of the test point.

We evaluate the performance of different masking methods on eye gaze estimation in a leave-one-out fashion, where each one of the eye images is used as the test data, the rest of the images are considered as training data, and the LLE weights are computed from the masked images. Figure 9 shows the average gaze estimation error $e$ (in terms of pixels in the projected screen) as a function of the lower dimension $m$ for the different linear embedding and masking algorithms, together with a baseline that employs the full-length original data. While MAPS algorithms again outperform other masking counterparts, there is a minor gap in performance between estimation from masked vs. full-length data. Furthermore, we believe that the improvement obtained by PCA vs. full-length data is due to the high level of noise observed in the image captures obtained with the low-power imaging architecture [20].

## VI. DISCUSSION AND FUTURE WORK

Our numerical experiments indicate that while each MAPS algorithm is well suited for a particular type of structure (local versus global), MAPS-Local often performs well when applied together with Isomap. We conjecture that this is due to the fact that MAPS-Local, by preserving the local structure, is also preserving the global structure that is relevant to Isomap.

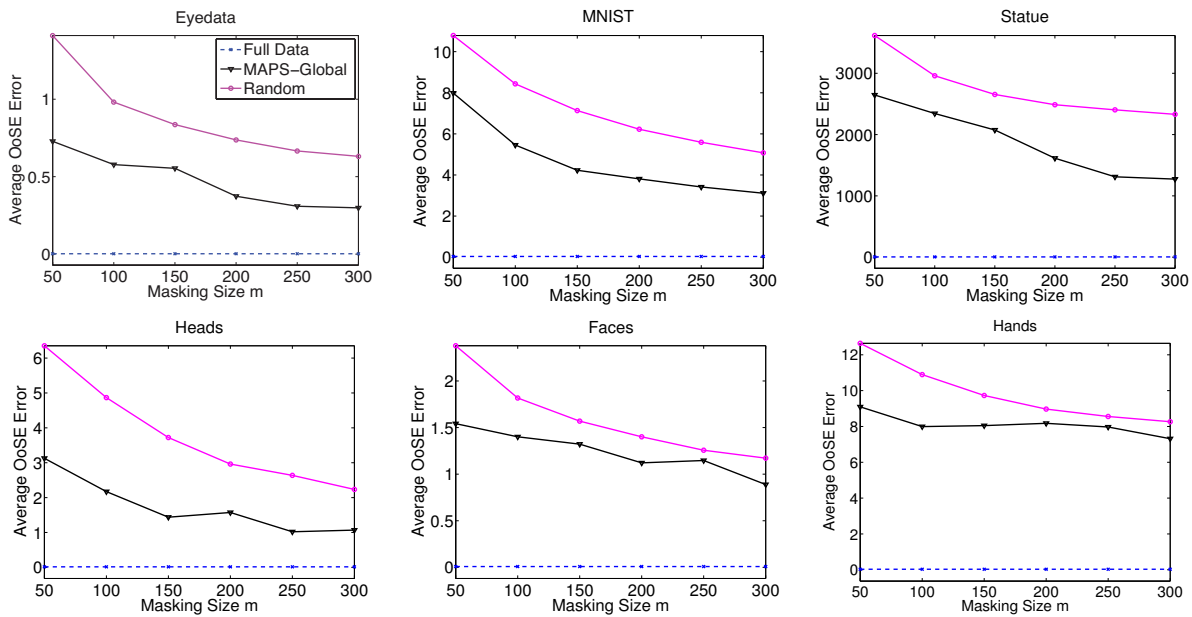

Fig. 7. Performance evaluation of Isomap OoSE for various datasets. The MAPS-Global algorithm consistently outperforms random masking.

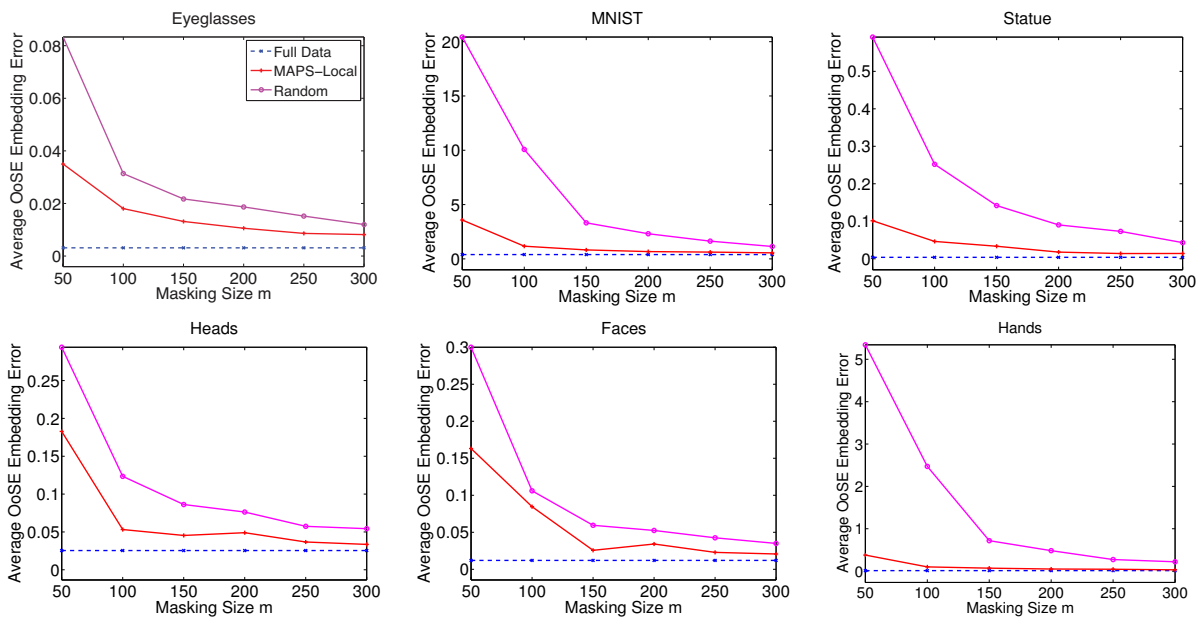

Fig. 8. Performance evaluation of LLE out-of-sample extension for various datasets. MAPS-Local consistently outperforms random masking.

In summary, if the activity in a set of images is localized spatially (which is often the case for image manifolds), then the proposed masking methods can identify subsets of pixels that successfully capture the relevant geometric structure of the dataset. As such, the degree to which the masking procedure (and hence the proposed masking algorithms) can preserve the relevant geometric structure in the data depends on how concentrated the activity is over a small subset of pixels. In addition, if the dataset's activity is spread out over all the pixels uniformly, then random masking would essentially be optimal, in which case the proposed MAPS algorithms would not do noticeably better than random masking. This can be observed from Faces and Heads datasets to some extent. In the latter datasets, random masking outperforms all the competing methods, except for the MAPS algorithms. Note that although in the latter two datasets the activity is spread out over a larger subset of pixels, compared to the other datasets, this spread is still not uniform over all the pixels of the image.

Since there are many other types of geometrical information leveraged by alternative manifold learning algorithms, it would be interesting to derive masking algorithms for them as well. Furthermore, there are several frameworks that can benefit from generalizations of the proposed masking algorithms. For instance, masking algorithms designed for datasets that are expressed as a union of manifolds can find applications in classification and pattern recognition. One may also leverage temporal information in video sequences to design more efficient manifold masking algorithms that take advantage of such temporal correlation.

On the connection between feature selection schemes and the proposed masking algorithms, note that the application of feature selection in supervised learning problems is driven by the goal of minimizing the estimation distortion or the regression/classification error, respectively. Our proposed manifold learning feature selection schemes are driven by the goal of minimizing the distortion of the embedding obtained via nonlinear manifold learning from the selected features vs. the embedding obtained from all features. For this purpose, we have derived data metrics that are specific to the geometric structure exploited by the considered manifold learning algorithms. The use of such a metric in place of the actual learning algorithm links our proposed approaches to the filter class of feature selection methods. One could derive alternative approaches to mask design by leveraging alternative feature selection schemes (such as backward or bidirectional elimination) similarly.

## VII. Conclusions

We have considered the problem of selecting image masks that aim to preserve the nonlinear manifold structure used in parameter estimation from images, in order to be able to learn the manifolds directly from the masked image data. Such a formulation enables a new form of compressive sensing using novel imaging sensors that feature power consumption growing with the number of pixels sensed. Our experimental evidence shows that the algorithms proposed for preservation of global and local geometric structure of the manifold outperform competing approaches, while requiring only a fraction of the computational cost. As a specific example, we have shown the potential of manifold learning from masked images for an eye gaze tracking application as an example application in cyber-physical systems.

## VIII. Acknowledgements

We thank Benjamin Marlin, Deepak Ganesan, and Addison Mayberry for helpful discussions, and for providing power consumption figures for the eyeglasses dataset.